\documentclass[twoside]{article}

\usepackage{arxiv}

\usepackage{lipsum}
\usepackage{amsmath}
\usepackage{amssymb}
\usepackage{bm}
\usepackage{enumitem}

\usepackage{setspace}

\usepackage{algorithm}
\usepackage{algpseudocode}

\usepackage[dvipsnames]{xcolor}
\newcommand{\RNum}[1]{\uppercase\expandafter{\romannumeral #1\relax}}

\usepackage{graphicx}
\usepackage{subfigure}

\usepackage[hidelinks]{hyperref}

\DeclareMathOperator*{\argmax}{arg\,max}

\usepackage[round]{natbib}

\begin{document}

\twocolumn[
\title{Efficient Bayesian Experimental Design for Implicit Models}
\author{Steven Kleinegesse \And Michael U. Gutmann}
\address{School of Informatics \\ University of Edinburgh\\  \href{mailto:steven.kleinegesse@ed.ac.uk}{steven.kleinegesse@ed.ac.uk} \And School of Informatics\\ University of Edinburgh\\ \href{mailto:michael.gutmann@ed.ac.uk}{michael.gutmann@ed.ac.uk}}]

\begin{abstract}
Bayesian experimental design involves the optimal allocation of resources in an experiment, with the aim of optimising cost and performance. For implicit models, where the likelihood is intractable but sampling from the model is possible, this task is particularly difficult and therefore largely unexplored. This is mainly due to technical difficulties associated with approximating posterior distributions and utility functions. We devise a novel experimental design framework for implicit models that improves upon previous work in two ways. First, we use the mutual information between parameters and data as the utility function, which has previously not been feasible. We achieve this by utilising Likelihood-Free Inference by Ratio Estimation (LFIRE) to approximate posterior distributions, instead of the traditional approximate Bayesian computation or synthetic likelihood methods. Secondly, we use Bayesian optimisation in order to solve the optimal design problem, as opposed to the typically used grid search or sampling-based methods. We find that this increases efficiency and allows us to consider higher design dimensions.
\end{abstract}

\section{INTRODUCTION}

In all scientific disciplines, performing experiments and therewith collecting data is an essential part to improving our understanding of the world around us.
It is, however, usually not trivial to decide where and how to collect the data; \emph{experimental design} is therefore concerned with the allocation of resources when conducting an experiment. The general aim is to find design features, or experimental configurations, that may improve parameter estimations or compare competing models.
In essence, the underlying question in experimental design is: where and how do we have to collect data in order to optimise cost and performance? For instance, in epidemiology we might be concerned about when to count the number of infected in a population. In this case, we could be trying to find the optimal measurement time that results in the most informative estimation of the disease model parameters.
Traditional experimental design uses frequentist approaches that are usually based on the Fisher information matrix~\citep[e.g.][]{Fedorov1972, Atkinson1992}; this is a well-established field. The frequentist framework however, does not work well for optimising non-linear problems, as only locally-optimal designs can be obtained~\citep{Ryan2016}.
Bayesian statistics has mature theory addressing this issue, but due to the computational costs involved, the field of \emph{Bayesian experimental design} has only recently become popular. Much of this high cost is incurred by computing the so-called \emph{expected utility} function $U(\mathbf{d})$ that is used to determine the optimal design $\mathbf{d}^\ast$ (e.g.\ the optimal measurement time). A popular and principled choice for this expected utility function is the \emph{mutual information} between parameters $\bm{\theta}$ and simulated data $\mathbf{y}$ at design $\mathbf{d}$~\citep[see][]{Ryan2016}.

There exists extensive work on Bayesian experimental design for \emph{explicit} models, where the likelihood is analytically known or can be easily computed~\citep[see][for a review]{Ryan2016}. There has, however, been little work on designing experiments for \emph{implicit} models, where the likelihood is intractable and the model is specified in terms of a stochastic data generating process or \emph{simulator}. These models are common in the natural sciences and appear in many disciplines; examples include epidemiology~\citep{Ricker1954, Numminen2013} and cosmology~\citep{Schafer2012, Alsing2018}. It is thus crucial to develop efficient methods for experimental design that apply to these models, which is the aim of this paper.

\paragraph{Previous Work}

Because the likelihood for implicit models is intractable, exact posterior computation is difficult. Likelihood-free inference methods have emerged to address this issue, with the most prevalent methods being Approximate Bayesian Computation (ABC)~\citep{Rubin1984} and Synthetic Likelihood (SL)~\citep{Wood2010}. Some of the earliest work of Bayesian experimental design for models with intractable likelihood was done by e.g.~\citet{Cook2008}, ~\citet{Liepe2013} and~\citet{Drovandi2013}; the latter was the first to use ABC rejection sampling~\citep{Beaumont2002} to obtain posterior samples for experimental design.
Most of the work that followed~\citep[e.g.][]{Dehideniya2018, Price2016,Hainy2016b} also used the same method.

Mutual information is typically the preferred choice for the expected utility function in Bayesian experimental design, as resulting optimal designs yield consistent and efficient parameter estimates~\citep{Paninski2005, Ryan2016}. For implicit models, however, computing the mutual information is hard, because of the difficulties associated with evaluating posterior densities. \citet{Drovandi2013} thus used the inverse of the posterior variance of ABC samples as the expected utility function, called 'Bayesian D-Optimality', in order to find the optimal design that minimises parameter uncertainty. \citet{Cook2008} used moment closure to approximate the mutual information, while ~\citet{Liepe2013} used ABC to approximate the mutual information for a restricted class of models and~\citet{Price2016} approximated the same with Kernel Density Estimation (KDE). 

Finding an efficient experimental design is often cast as an optimisation problem. The Bayesian optimal design $\mathbf{d}^\ast$ that we are trying to find is the design point that maximises the expected utility function $U(\mathbf{d})$ over the whole design space,
\begin{equation} \label{eq:optdesign}
\mathbf{d}^\ast = \argmax_{\mathbf{d}} U(\mathbf{d}).
\end{equation}
For implicit models however, $U(\mathbf{d})$ is not available analytically in closed form and also expensive to evaluate; in addition, we also do not have access to its gradients.
Previous work on Bayesian experimental design, considering both explicit and implicit models, predominantly pertains to solving the above optimisation problem either by the sampling-based algorithm of~\citet{Muller99} or by grid search. The former has been found to converge slowly~\citep{Drovandi2013}, as finding the maximum of a set of samples is not easy. The latter approach is much more common but becomes unfeasible for high design dimensions, due to the extremely high amount of $\mathbf{U}(\mathbf{d})$ evaluations needed. There has been some work on improving this part of the experimental design process. Evolutionary algorithms were proposed~\citep{Price2018} and, in the context of explicit models,~\citet{Overstall2017} reduced the multivariate optimisation problem to a sequence of one-dimensional problems that were solved using Gaussian Processes (GP). 
This method was recently applied to the likelihood-free setting by~\citet{Overstall2018}.

\paragraph{Contributions}

In this paper, we propose an efficient Bayesian experimental design framework for implicit models that addresses the aforementioned technical difficulties.

\begin{enumerate}
\item We use the mutual information between model parameters and data as the utility function to find the optimal design. We achieve this by using Likelihood-Free Inference by Ratio Estimation (LFIRE)~\citep{Thomas2016}, instead of the traditional ABC or SL approaches, to approximate the posterior distribution for implicit models. 
\item We show that the optimisation problem in Equation~\ref{eq:optdesign} can be succesfully solved by means of Bayesian optimisation~\citep[e.g.][]{Shahriari2016}. This makes Bayesian experimental design in the likelihood-free setting more efficient and allows us to consider higher design dimensions.
\end{enumerate}
The remainder of the paper is structured as follows. Our novel design framework is explained in Section~\ref{sec:prop}. In Section~\ref{sec:exp} we test the performance of our framework on two epidemiological models and discuss the results. We then summarise our findings in Section~\ref{sec:concl}.

\section{PROPOSED METHOD} \label{sec:prop}

At its core, Bayesian experimental design requires us to compute an expected utility function $U(\mathbf{d})$ that describes the value of design $\mathbf{d}$ in learning about $\bm{\theta}$.
We then need to maximise this in order to find the optimal design point $\mathbf{d}^\ast$. We explain here how we address these two non-trivial steps efficiently for implicit models.

\subsection{Computing the Utility}

The choice of expected utility function strongly dictates the optimal designs that are found. Here we consider information-based utilities and focus on the computation of mutual information for implicit models. We consider the mutual information $\mathrm{MI}(\bm{\theta}, \mathbf{y} \mid \mathbf{d})$ between the model parameters $\bm{\theta}$ and the simulated data $\mathbf{y}$, conditioned on the design $\mathbf{d}$. This gives us a measure of the non-linear correlation between $\bm{\theta}$ and $\mathbf{y}$, i.e.~it tells us how 'much' we can learn about the model parameters given the data. This mutual information can also be expressed as the expected Kullback-Leibler Divergence $\mathrm{D_{KL}}$~\citep{Kullback1951} between the posterior distribution $p(\bm{\theta} \mid \mathbf{d}, \mathbf{y})$ and prior distribution $p(\bm{\theta})$~\citep{Ryan2016}.
In this phrasing, the optimal design $\mathbf{d}^\ast$ can be understood as the design that, on average, yields the largest information gain about the parameters when observing the data. The expected utility that we need to maximise is then
\begin{align}
U(\mathbf{d}) &= \mathrm{MI}(\bm{\theta}, \mathbf{y} \mid \mathbf{d}) \label{eq:mutual1} \\
&= \mathbb{E}_{p(\mathbf{y} \mid \mathbf{d})}[\mathrm{D_{KL}}(p(\bm{\theta} \!\mid\! \mathbf{d}, \mathbf{y}) \mid\mid p(\bm{\theta}))] \label{eq:exputility} \\
&= \int \log\left[\frac{p(\bm{\theta} \!\mid\! \mathbf{d}, \mathbf{y})}{p(\bm{\theta})}\right] p(\mathbf{y} \!\mid\! \bm{\theta}, \mathbf{d}) p(\bm{\theta}) \mathrm{d}\bm{\theta} \mathrm{d}\mathbf{y}, \label{eq:mutual3}
\end{align}
see e.g.\ \citet{Ryan2016}. We here made the typical assumption that $p(\bm{\theta} \mid \mathbf{d})$ = $p(\bm{\theta})$. 

The integral in Equation~\ref{eq:mutual3} is generally high-dimensional and a standard way of approximating it is by means of Monte-Carlo integration, i.e.~
\begin{align}
U(\mathbf{d}) &\approx \frac{1}{N} \sum_{i=1}^N \log\left[\frac{p(\bm{\theta}^{(i)} \!\mid\! \mathbf{d}, \mathbf{y}^{(i)})}{p(\bm{\theta}^{(i)})}\right], \label{eq:mutual_mc}
\end{align}
where $\mathbf{y}^{(i)} \sim p(\mathbf{y} \mid \mathbf{d}, \bm{\theta}^{(i)})$ and $\bm{\theta}^{(i)} \sim p(\bm{\theta})$. This approximation requires samples from the prior distribution, corresponding samples from the data generating distribution and density evaluations of the posterior and prior distribution.

For implicit models, the computation of the posterior distribution in Equation~\ref{eq:mutual_mc} is hard. Using the traditional ABC method, we would only obtain posterior samples but not posterior densities, making the computation of Equation~\ref{eq:mutual_mc} even more difficult.
Ratio estimation approaches on the other hand can yield ratios $r(\mathbf{d}, \mathbf{y}, \bm{\theta})$ of the likelihood to the marginal and therefore, by Bayes' Rule, also yield the ratios of the posterior density to prior density, i.e.~
\begin{equation} \label{eq:ratio}
r(\mathbf{d}, \mathbf{y}, \bm{\theta}) = \frac{p(\mathbf{y} \mid \bm{\theta}, \mathbf{d})}{p(\mathbf{y} \mid \mathbf{d})} = \frac{p(\bm{\theta} \mid \mathbf{d}, \mathbf{y})}{p(\bm{\theta})},
\end{equation}
which is the intractable ratio in Equation~\ref{eq:mutual_mc}. An overview of methods for ratio estimation methods can be found in~\citet{Sugiyama2012}. We here use the Likelihood-Free Inference by Ratio Estimation (LFIRE) framework of~\citet{Thomas2016} to approximate the above ratio. Importantly, the approximated ratios can be used to estimate both the posterior and the mutual information.
The LFIRE ratio is approximated by solving a logistic regression problem between data simulated from $p(\mathbf{y} \mid \bm{\theta}, \mathbf{d})$ and data simulated from $p(\mathbf{y} \mid \mathbf{d})$.
We shall omit further details here and direct the reader to the work of \citet{Thomas2016} for more information. Throughout this work, we used the same settings as them and, in particular, we used $1{,}000$ data points from each of the two distributions. 

By substituting the LFIRE ratio $r(\mathbf{d}, \mathbf{y}, \bm{\theta})$ into the logarithm in Equation~\ref{eq:mutual_mc}, we can approximate the mutual information in a straightforward way, without having to explicitly compute posterior and prior densities, i.e.~
\begin{align}
U(\mathbf{d}) &\approx \frac{1}{N} \sum_{i=1}^N \log\left[r(\mathbf{d}, \mathbf{y}^{(i)}, \bm{\theta}^{(i)})\right], \label{eq:mutual_final}
\end{align}
where $\mathbf{y}^{(i)} \sim p(\mathbf{y} \mid \mathbf{d}, \bm{\theta}^{(i)})$ and $\bm{\theta}^{(i)} \sim p(\bm{\theta})$. In other words, for a given design $\mathbf{d}$ we first need to obtain prior samples $\{\bm{\theta}^{(i)}\}_{i=1}^{N}$ and then use them to simulate data points $\{\mathbf{y}^{(i)}\}_{i=1}^{N}$. These pairs of prior samples and data samples are then used to compute $N$ ratios $\{r(\mathbf{d}, \mathbf{y}^{(i)}, \bm{\theta}^{(i)})\}_{i=1}^{N}$, allowing us to approximate the expected utility according to Equation~\ref{eq:mutual_final}. Assuming that we choose a prior that is easy to sample from and that the process of generating data from the implicit model is not overly expensive, most of the computational cost lies in the LFIRE ratio computations.
We summarise the computation of the mutual information for implicit models in Algorithm~\ref{algo:mutual}.
\begin{algorithm}
\caption{Mutual Information Computation via LFIRE Ratios}\label{algo:mutual}
\begin{algorithmic}[1]
\State {Sample from the prior: $\bm{\theta}^{(i)} \sim p(\bm{\theta})$ for $i=1, \dots, N$}
\For {i=1 to i=N}
    \State {Simulate data: $\mathbf{y}^{(i)} \sim p(\mathbf{y} \mid \mathbf{d}, \bm{\theta}^{(i)})$}
    \State {Compute the  ratio $r(\mathbf{d}, \mathbf{y}^{(i)}, \bm{\theta}^{(i)})$} by LFIRE 
\EndFor
\State {Compute $U(\mathbf{d}) \approx \frac{1}{N} \sum_{i=1}^N \log\left[r(\mathbf{d}, \mathbf{y}^{(i)}, \bm{\theta}^{(i)})\right]$}
\end{algorithmic}
\end{algorithm}

\subsection{Optimising the Utility}

Evaluating the expected utility is costly and gradients are not readily available. We thus propose to use Bayesian optimisation (BO)~\citep{Shahriari2016} to solve the optimisation problem in Equation~\ref{eq:optdesign}. BO is a popular optimisation scheme for objective functions that we can evaluate but whose form and gradients are unknown, or expensive to evaluate, and hence well-suited for Bayesian experimental design for implicit models.

The general idea of BO is to build a probabilistic model of the objective function and then use an acquisition function to decide where to evaluate it next. In our case, the objective function is the expected utility in Equation~\ref{eq:mutual_final}. For the probabilistic model we use Gaussian Processes (GP)~\citep{Rasmussen2005} and for the acquisition function we use Expected Improvement (EI)~\citep{Mockus1978}. These are both popular and well-tested choices. We use a Gaussian kernel for the GP model, but for design dimensions higher than 10 there exist more scalable kernels~\citep[e.g.][]{Minasny2005, ChangYong2018}. For the BO stage of our design framework we use the GPyOpt package in Python~\citep{gpyopt2016}. In addition to being practical for expensive evaluations, BO smoothes out noise introduced by the Monte-Carlo approximation and is therefore likely to also improve the estimate of $U(\mathbf{d})$.

\subsection{Obtaining the Posterior} \label{sec:samples}

After having found an optimal design $\mathbf{d}^\ast$ that maximises the expected utility $U(\mathbf{d})$, we can make a real-world observation $\mathbf{y}^\ast$. LFIRE allows us to use the already computed ratios at $\mathbf{d}^\ast$ to estimate the posterior of $\bm{\theta}$ given $\mathbf{y}^\ast$ and obtain samples from it. By rearranging Equation~\ref{eq:ratio} we can easily compute the posterior density at a certain model parameter, given that we can evaluate the prior density. While other approaches are possible, to obtain posterior samples via the LFIRE method, we define weights $w^{(i)}$ for every prior sample $\bm{\theta}^{(i)}$; each weight is the LFIRE ratio evaluated at the real-world observation $\{\mathbf{d}^\ast, \mathbf{y}^\ast\}$, i.e.~$w^{(i)} = r(\mathbf{d}^\ast, \mathbf{y}^\ast, \bm{\theta}^{(i)})$. After normalising the weights, i.e.~$W^{(i)} = w^{(i)} / \sum w^{(i)}$, we then resample from the set of prior parameters $\{\bm{\theta}^N_{i=1}\}$ according to the categorical distribution $\mathrm{cat}(\{W^{(i)}\}^N _{i=1})$.
\section{EXPERIMENTS} \label{sec:exp}

In this section, we test our novel design framework on two implicit models from epidemiology, the Death Model~\citep{Cook2008} and the SIR Model~\citep{Allen2008}. The former has a tractable likelihood in closed form, allowing us to compare approximations to an analytical solution, while the latter does not. For both models the design variable $\mathbf{d}$ is time and therefore the aim is to find out at what times $\tau$ we should make measurements in order to most accurately estimate the model parameters.

\subsection{Example Implicit Models}

\paragraph{Death Model}

The Death Model is a stochastic process that describes the decline of a population due to some infection. The change from a susceptible state $S$ to an infected state $I$ is given by a continuous-time Markov process that we discretised. The process is parametrised by an infection rate $b$ and at any time $t$, each susceptible individual has a chance $p_{\mathrm{inf}}(t) = 1 - \exp(-bt)$ of getting infected~\citep{Cook2008}. At a particular time $t$, the total number of individuals $\Delta I (t)$ moving from state $S$ to state $I$ is given by a sample from a Binomial distribution~\citep{Cook2008}, i.e.~
\begin{align}
\Delta I (t) \sim \mathrm{Bin}(\Delta I (t); N - I(t), p_{\mathrm{inf}}(\Delta t)), \label{eq:death1}
\end{align}
where $N$ is the invariant total population, $\Delta t$ is the step size, set to $0.01$ throughout, and we choose that $I(t = 0) = 0$. As a time series, the number of infected is then given by $I(t+\Delta t) = I(t) + \Delta I (t)$.

Let $\tau_1, \ldots, \tau_n$ be the measurement times at which we observe the number of infected. The likelihood for the Death Model is analytically tractable~\citep{Cook2008}. For $n$ observations $\{\tau_k, S(\tau_k)\}_{k=1}^n$ of the number of susceptibles, given by $S(\tau_k) = N - I(\tau_k)$, and a model parameter $b$, we obtain $p(\{S(\tau_k)\}_{k=1}^n \mid b) = \prod_{k=1}^n \mathrm{Bin}(S(\tau_k); S(\tau_{k-1}), \mathrm{exp}(-b(\tau_k - \tau_{k-1})))$, where $\tau_0 = 0$ and $S(\tau_0) = N$~\citep{Cook2008}. Using this expression for the likelihood we can then obtain a posterior distribution by Bayes' Rule. This enables us to compute the expected utility in Equation~\ref{eq:exputility} and compare it to the LFIRE approximation.

\paragraph{SIR Model}

The SIR Model~\citep{Allen2008} is a more complex version of the Death Model where, in addition to the number of susceptibles $S(t)$ and infected $I(t)$, we also have a recovered population $R(t)$ that cannot be further infected.

We define the probability of a susceptible getting infected as $p_{\mathrm{inf}}(t) = \beta I(t)/N$, where $\beta \in [0,1]$. Similarly, the probability of an infected recovering from the disease is $p_{\mathrm{rec}}(t) = \gamma$, where $\gamma \in [0,1]$. At a particular time $t$, let the number of susceptibles that get infected be $\Delta I (t)$ and the number of infected that recover be $\Delta R (t)$; these two population changes are computed by sampling from a Binomial distribution,
\begin{align} \label{eq:popchange}
\Delta I (t) &\sim \text{Bin}(S(t), p_{\text{inf}}(t)) \\
\Delta R (t) &\sim \text{Bin}(I(t), p_{\text{rec}}(t)).
\end{align}
This results in an unobserved time-series of $S$, $I$ and $R$ by doing the following updates:
\begin{align} \label{eq:statechange}
S(t+\Delta t) &= S(t) - \Delta I (t) \\
I(t+\Delta t) &= I(t) + \Delta I (t) - \Delta R (t)\\
R(t+\Delta t) &= R(t) + \Delta R (t).
\end{align}
We shall start this time series with $N-1$ susceptibles, one infected, zero recovered and use a discrete time step of $\Delta t = 0.01$ throughout. The actual time at which we do measurements is again given by $\tau$, resulting in a single data point $(S(\tau), I(\tau), R(\tau))$.

\subsection{Death Model Results}

The aim for the Death Model is to estimate the infection rate $b$ as efficiently as possible.

\paragraph{One-Dimensional Designs}

We put a truncated Gaussian prior of mean one and variance one over the model parameter $b$, such that $b > 0$, and sample $1,000$ prior parameters $\{b^{(i)}\}^{1000}_{i=1}$ from it. The design space covers the range $0 < \tau \leq 4$ and, when optimising the expected utility $U(\tau)$ via grid search, we choose grid sizes of $\Delta \tau = 0.1$.

\begin{figure}[!ht]
\begin{center}
\centerline{\includegraphics[width=\columnwidth]{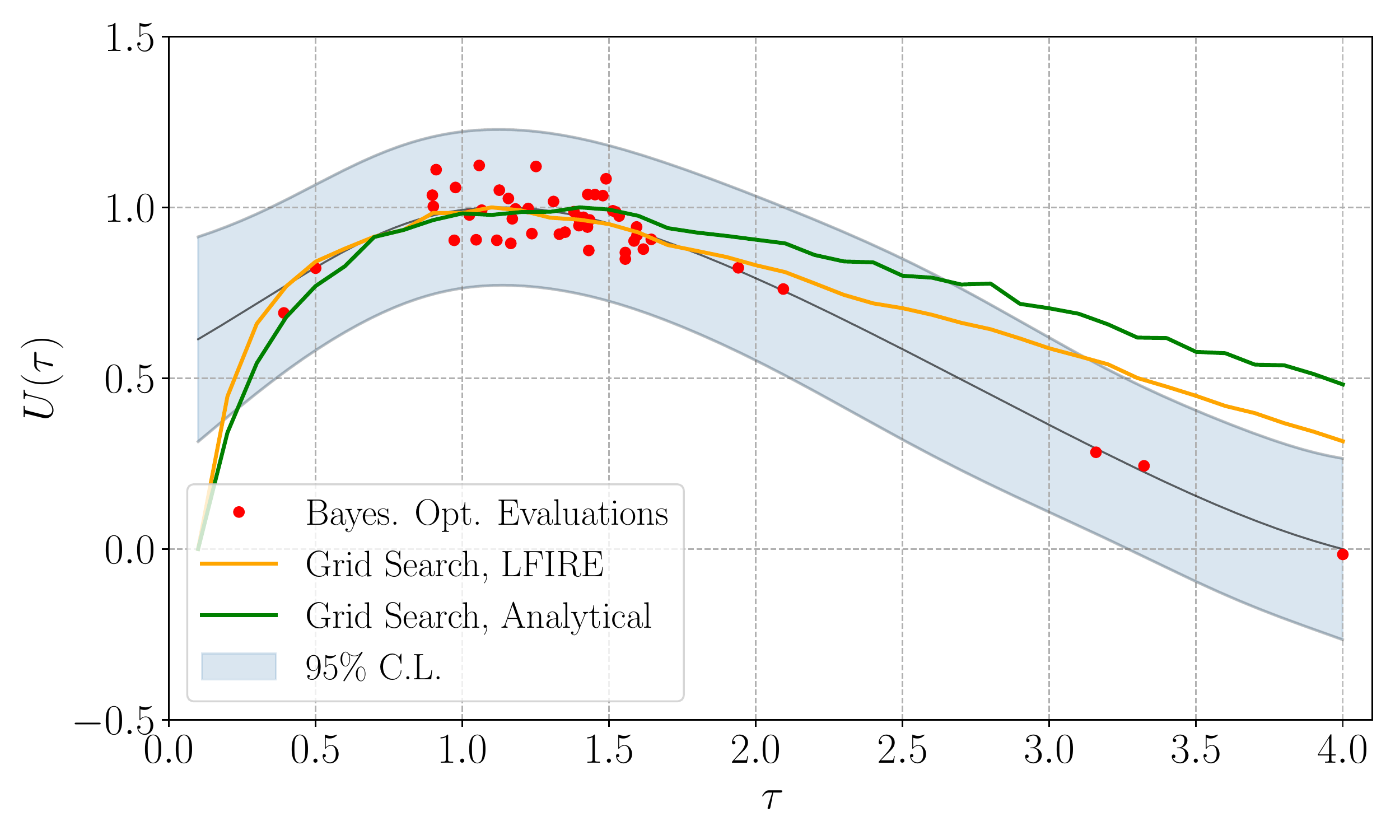}}
\caption[Filler]{Expected utilities $U(\tau)$ of the Death Model for the grid search method and the BO method. Also shown is the analytic expected utility for the grid search method. Curves are normalised to be in $(0,1)$.}
\label{fig:death_comparison}
\end{center}
\vskip -5mm
\end{figure}

We then optimise the expected utility function by Bayesian optimisation (BO), according to the framework outlined in Section~\ref{sec:prop}. We compare our method to optimising $U(\tau)$ by grid search, for both the LFIRE approximation and the analytic computation of the expected utility. These expected utilities are shown in Figure~\ref{fig:death_comparison}.

The expected utility approximated with LFIRE ratios closely matches the analytical expected utility around its peak while decaying more quickly for large $\tau$. This justifies using the LFIRE approximation in cases where we cannot compute the mutual information exactly, such as for the SIR Model. The BO method results in a similar expected utility as the grid search method. Unlike the grid search method however, Bayesian optimisation results in few evaluations where $U(\tau)$ is low, focusing more on regions where it is high and thus yielding a higher resolution in the peak region. This results in some large discrepancies between the grid search method and Bayesian optimisation method away from the peak region, i.e.~at the boundaries. The optimal design times $\tau^\ast$ and corresponding $U(\tau^\ast)$ values, computed by using the analytic likelihood, are: $(\tau^\ast, U(\tau^\ast)) = (1.40,1.347)$ for the grid search method with analytic computations, $(\tau^\ast, U(\tau^\ast)) = (1.10,1.350)$ for the grid search method with LFIRE approximations and $(\tau^\ast, U(\tau^\ast)) = (1.06,1.359)$ for the BO method with the LFIRE approximations. The optimal design time for the analytic computation is slightly larger than for the two methods using LFIRE approximations. The actual expected utility values, however, are all close together. This means that, for the Death Model, there is a range of optimal design times that result in comparable utility values, which is reflected by the flat peak in Figure~\ref{fig:death_comparison}.

\begin{figure}[!ht]
\begin{center}
\centerline{\includegraphics[width=\columnwidth]{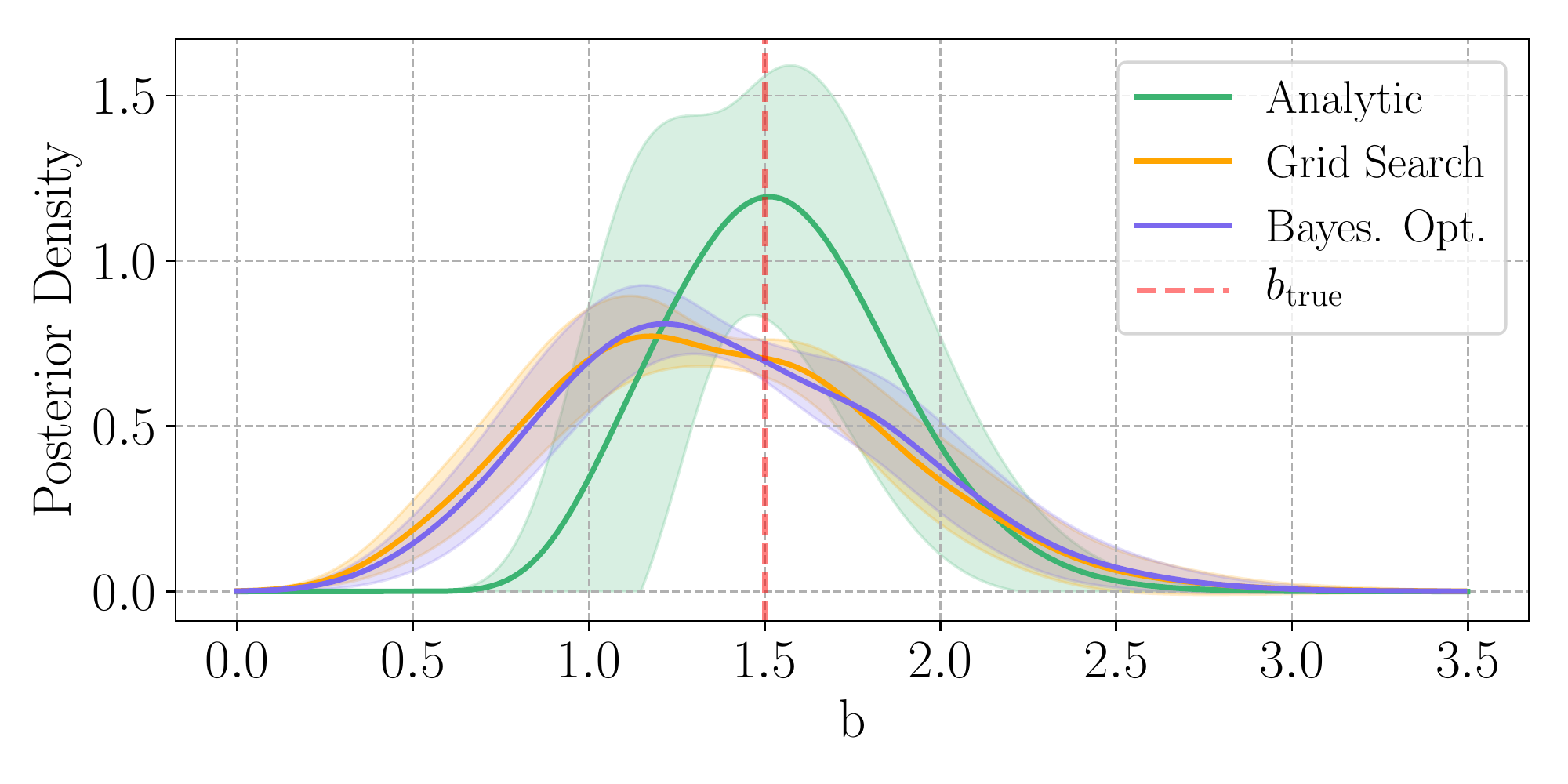}}
\caption[Filler]{Comparison of the Death Model mean posterior densities for different methods. The shaded areas indicate one standard deviation.}
\label{fig:death_post}
\end{center}
\vskip -5mm
\end{figure}

Using $b_{\mathrm{true}}=1.5$ as the model parameter to generate 'real-world' observations $I^\ast$ at the optimal times $\tau^\ast$, we obtain $10,000$ LFIRE posterior samples for the grid search and BO methods, according to the procedure outlined in Section~\ref{sec:samples}. We then apply Gaussian kernel density estimation (KDE) to these samples to smooth out the resulting posterior density. Doing this for $50$ real-world observations $I^\ast$ allows us to obtain $50$ posterior densities reflecting the possible variation in the data measured at time $\tau^\ast$. For comparison, we similarly compute the exact posterior distribution by using the tractable likelihood function mentioned previously. The mean of the posterior densities and their standard deviations are shown in Figure~\ref{fig:death_post}.

The grid search and BO method result in extremely similar posterior densities. The exact posterior density is narrower than the approximations, reflecting the approximation error of the LFIRE approach. Even though these approximate posterior distributions have a discrepancy to the exact posterior, the corresponding $U(\tau)$ functions are still similar (see Figure~\ref{fig:death_comparison}). This indicates that, on average, the divergence between individual posterior and prior distributions is still comparable for the different methods (see Equation~\ref{eq:exputility}).
Using these mean posterior densities, we find the median model parameters to be: $\widehat{b}_{\mathrm{grid}}=1.34$, $\widehat{b}_{\mathrm{BO}}=1.36$ and $\widehat{b}_{\mathrm{an}}=1.53$ for the grid search, BO and analytic method, with 95\% credibility intervals for $b$ equal to $(0.50,2.40)$, $(0.54,2.41)$ and $(0.96,2.24)$, respectively.

\begin{figure}[!ht]
\begin{center}
\centerline{\includegraphics[width=\columnwidth]{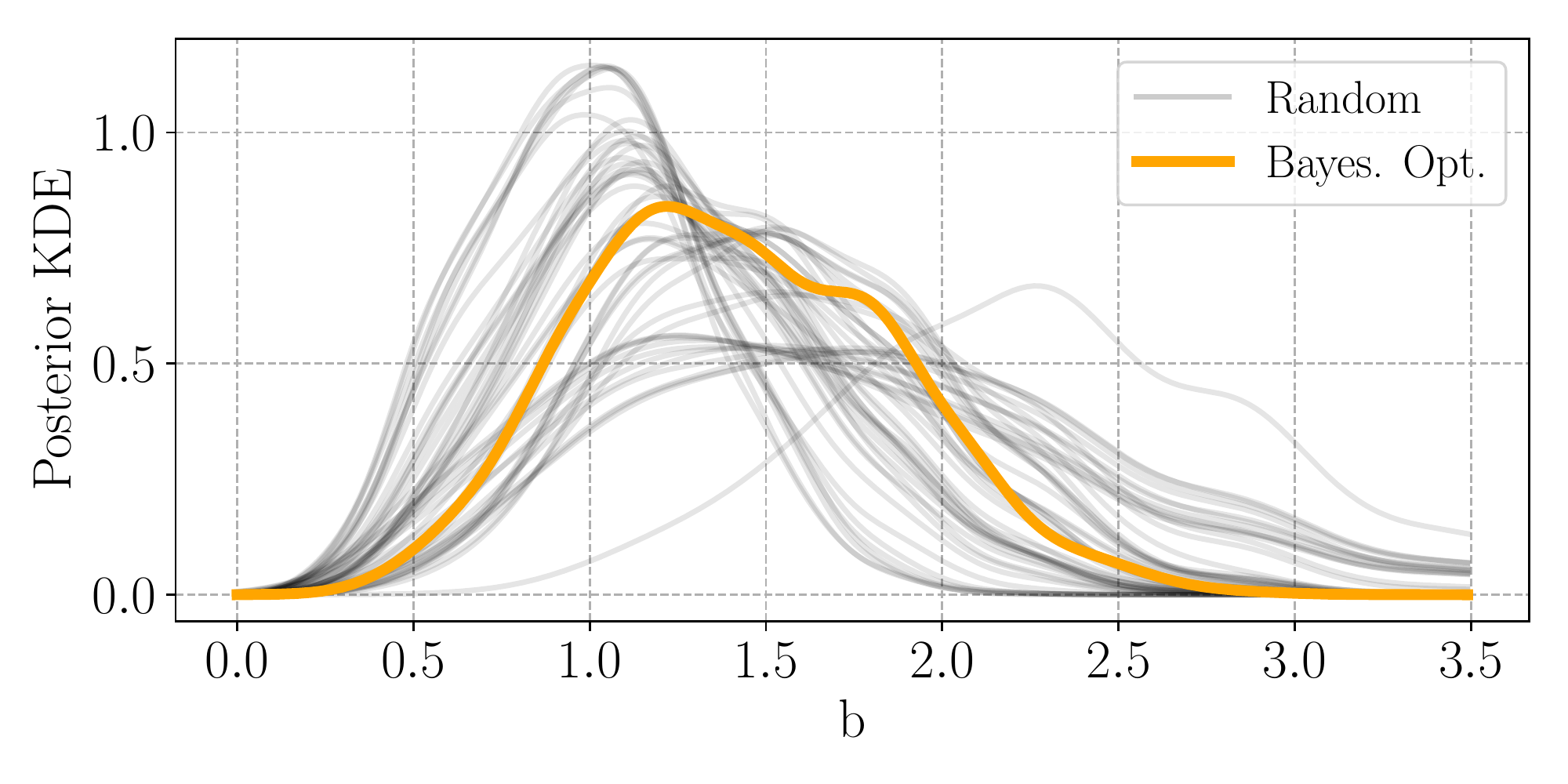}}
\caption[Filler]{Death Model posterior densities for the baseline of randomly selecting design points, together with the BO method mean posterior density.}
\label{fig:death_random}
\end{center}
\vskip -5mm
\end{figure}

At this point we would like to emphasise the importance of designing an experiment, as opposed to just randomly selecting an experimental design. To do so, we compute a baseline where we randomly select an optimal design $\tau^{\ast}_{\mathrm{ran}}$ from the design space. Using this random design point and a corresponding real-world observation $I^{\ast}_{\mathrm{ran}}$, we compute LFIRE ratios, as done previously. These are then used to compute a resulting posterior distribution, which is again smoothed out by Gaussian KDE. Because of the inherent randomness, we do this several times and compute a set of posterior densities. In Figure~\ref{fig:death_random} we compare $50$ of these baseline posterior densities to the mean posterior density obtained via the BO method.

As can be seen from Figure~\ref{fig:death_random}, there is much fluctuation involved in randomly selecting optimal designs. If the experimenter is unlucky and selects a design point that is highly unfavourable, e.g.~see $\tau=4$ in Figure~\ref{fig:death_comparison}, then the resulting posterior distribution is wide and the parameter estimate uncertain. The large variety in posterior distributions largely motivates the use of Bayesian experimental design in general.

\paragraph{High-Dimensional Designs}

So far, we have only considered a one-dimensional design variable, the measurement time $\tau$. We can, however, increase the design dimensions of this problem by rephrasing the premise. We shall now consider the problem of selecting $n$ optimal design times, instead of just one; this is referred to as \emph{non-myopic} Bayesian experimental design. To do this, our design variable becomes a $n$-dimensional vector, $\mathbf{d} = [\tau_1, \tau_2, \dots, \tau_n]^\top$. We naturally add the constraint that time must be ordered, i.e.~$\tau_1 < \tau_2 < \dots < \tau_n$, and then sequentially compute the number of infected at each design time to build the data vector $\mathbf{y} = [I(\tau_1), I(\tau_2), \dots, I(\tau_n)]^\top$. The values of this data vector depend on each other according to Equation~\ref{eq:death1}, i.e.~$I(\tau_n)$ depends on $I(\tau_{n-1})$ and so on. The computation of the LFIRE ratios is then done as before, with the difference that we have an $n$-dimensional design variable $\mathbf{d}$ and an $n$-dimensional data vector $\mathbf{y}$; the expected utility is also computed as previously, according to Algorithm~\ref{algo:mutual}.

\begin{figure}[!ht]%
\centering
\subfigure[Convergence]{%
\label{fig:multideath_first}%
\includegraphics[height=1.45in]{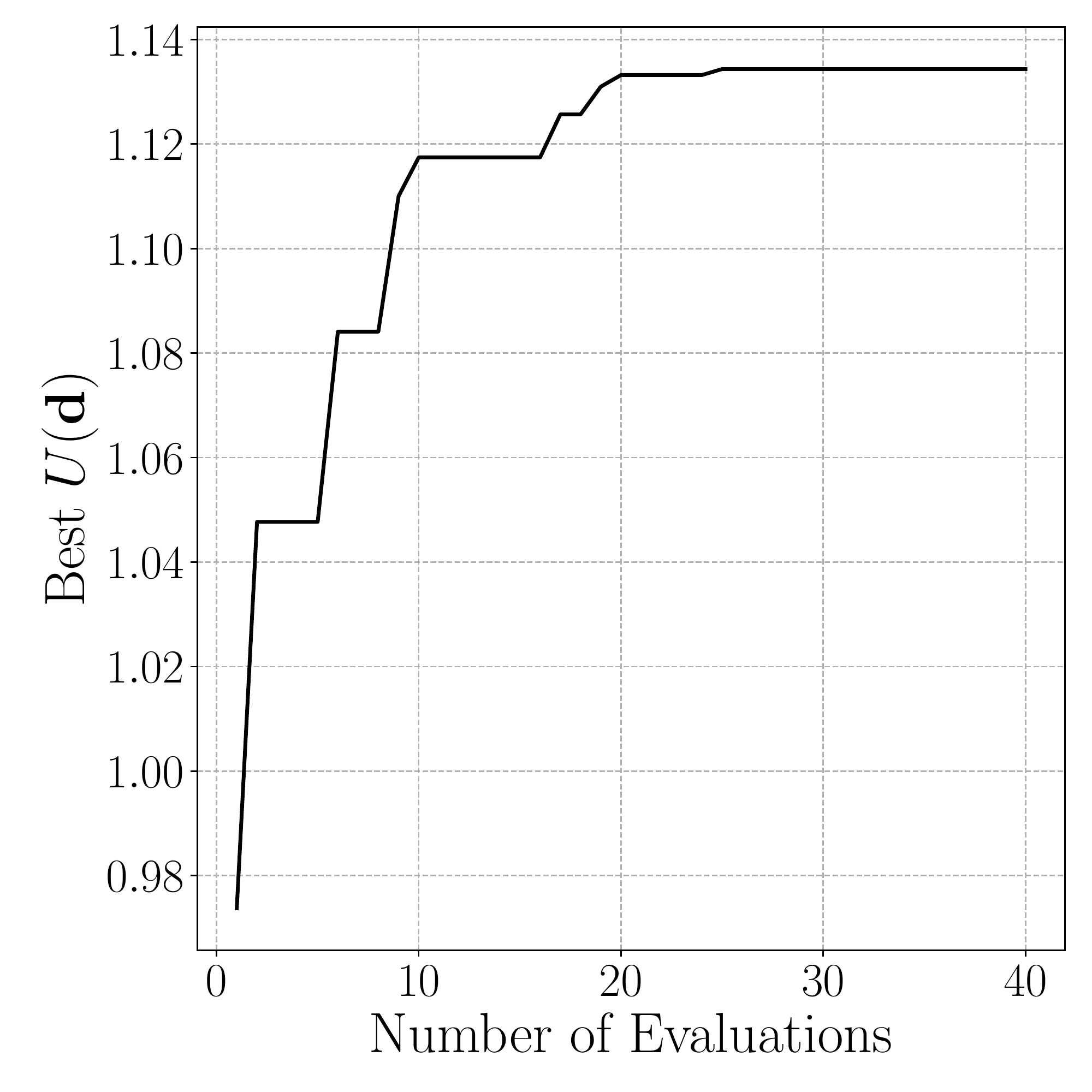}}%
\qquad
\subfigure[Posterior]{%
\label{fig:multideath_second}%
\includegraphics[height=1.45in]{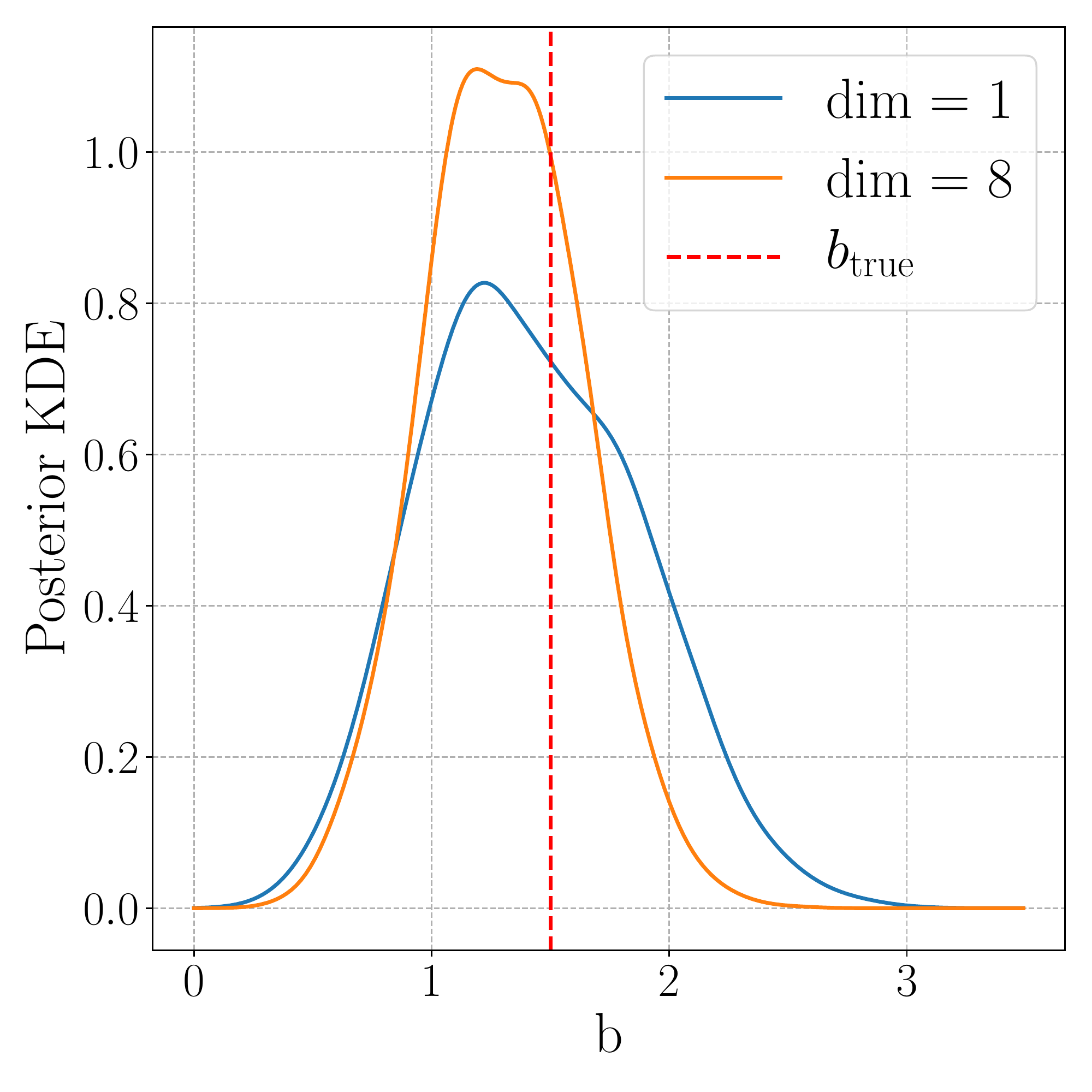}}%
\caption{(a) Convergence of the best expected utility $U(\mathbf{d})$ value as a function of number of evaluations for the Death Model with $n=8$ design dimensions. (b) The posterior distribution, smoothed out by Gaussian KDE, corresponding to the optimal design $\mathbf{d}^\ast$. Also shown is the posterior distribution for $n=1$.}
\vskip -5mm
\end{figure}

It is computationally unfeasible to perform grid search in high design dimensions, as the number of $U(\mathbf{d})$ evaluations required increases dramatically. Optimising $U(\mathbf{d})$ via Bayesian optimisation (BO) allows us to decrease the computational cost, as we only need to explore the design space where the expected utility is potentially high, as was the case in Figure~\ref{fig:death_comparison}. In addition, because we have more evaluations in the peak regions of $U(\mathbf{d})$ and we smooth out the noise introduced by the Monte-Carlo approximation in Algorithm~\ref{algo:mutual}, we increase the accuracy of our optimum estimate $\mathbf{d}^\ast$.

As a proof of concept, we consider a non-myopic design problem where the number of design dimensions is $n=8$. In other words, knowing that we can do eight experiments, we want to find out at what times we should take these measurements. Because of the increased dimensions, it is impossible to show the expected utility surface in the same way that we did in Figure~\ref{fig:death_comparison}. We again run the procedure outlined in Section~\ref{sec:prop} in order to find optimal measurement times $\mathbf{d}^\ast$. In Figure~\ref{fig:multideath_first} we show the convergence towards the optimum as a function of $U(\mathbf{d})$ evaluations. From this figure we find that we can converge to an optimum after around $20$ evaluations. If we had defined a four-dimensional grid instead with 40 points per dimensions, like we did for the one-dimensional situation in Figure~\ref{fig:death_comparison}, we would have had to do $76,904,685$ evaluations. It becomes apparent that we can drastically improve computational efficiency when using BO.

Using the procedure explained in Section~\ref{sec:prop}, we obtain $10,000$ LFIRE posterior samples corresponding to the optimal design $\mathbf{d}^\ast$ and then smooth out the posterior samples with Gaussian kernel density estimation (KDE). The resulting posterior density is shown in Figure~\ref{fig:multideath_second}, together with the posterior density for $n=1$. Expectedly, the posterior is narrower for $n=8$ than for $n=1$, due to having more data. The corresponding median estimate of the model parameter is $\widehat{b} = 1.29$ and the 95\% credibility interval $(0.68, 1.95)$.

As a comparison, we also compute the expected utility value at a design point $\mathbf{d}_{\mathrm{eq}}$ consisting of $8$ equidistant times; this might be an intuitive choice for an experimenter having no prior information. Using Algorithm~\ref{algo:mutual}, we obtain $U(\mathbf{d}_{\mathrm{eq}}) = 1.21$ for the equidistant design times and $U(\mathbf{d}^\ast) = 1.27$ for the optimal design times. Thus, unlike for $n=1$ seen before, if we can take many measurements, we do not get much improvement when designing a non-myopic experiment. This is natural to occur, in particular for the relatively simple Death model.

\subsection{SIR Model Results}

The aim for the SIR Model is to estimate the rate of infection $\beta$ and the rate of recovery $\gamma$ as efficiently as possible.

\paragraph{One-Dimensional Designs}

We put a uniform prior $\mathrm{U}(0,0.5)$ on both model parameters and sampled $1,000$ parameters $\{\beta^{(i)}, \gamma^{(i)}\}_{i=1}^{1000}$ from the prior. The data used in the LFIRE computations is $\{S(\tau), I(\tau), R(\tau)\}$ at a particular design time $\tau$. The design space covers the range $0 < \tau \leq 3$ and, when optimising the expected utility $U(\tau)$ via grid search, we choose step sizes of $\Delta \tau = 0.1$.

\begin{figure}[!ht]
\begin{center}
\centerline{\includegraphics[width=\columnwidth]{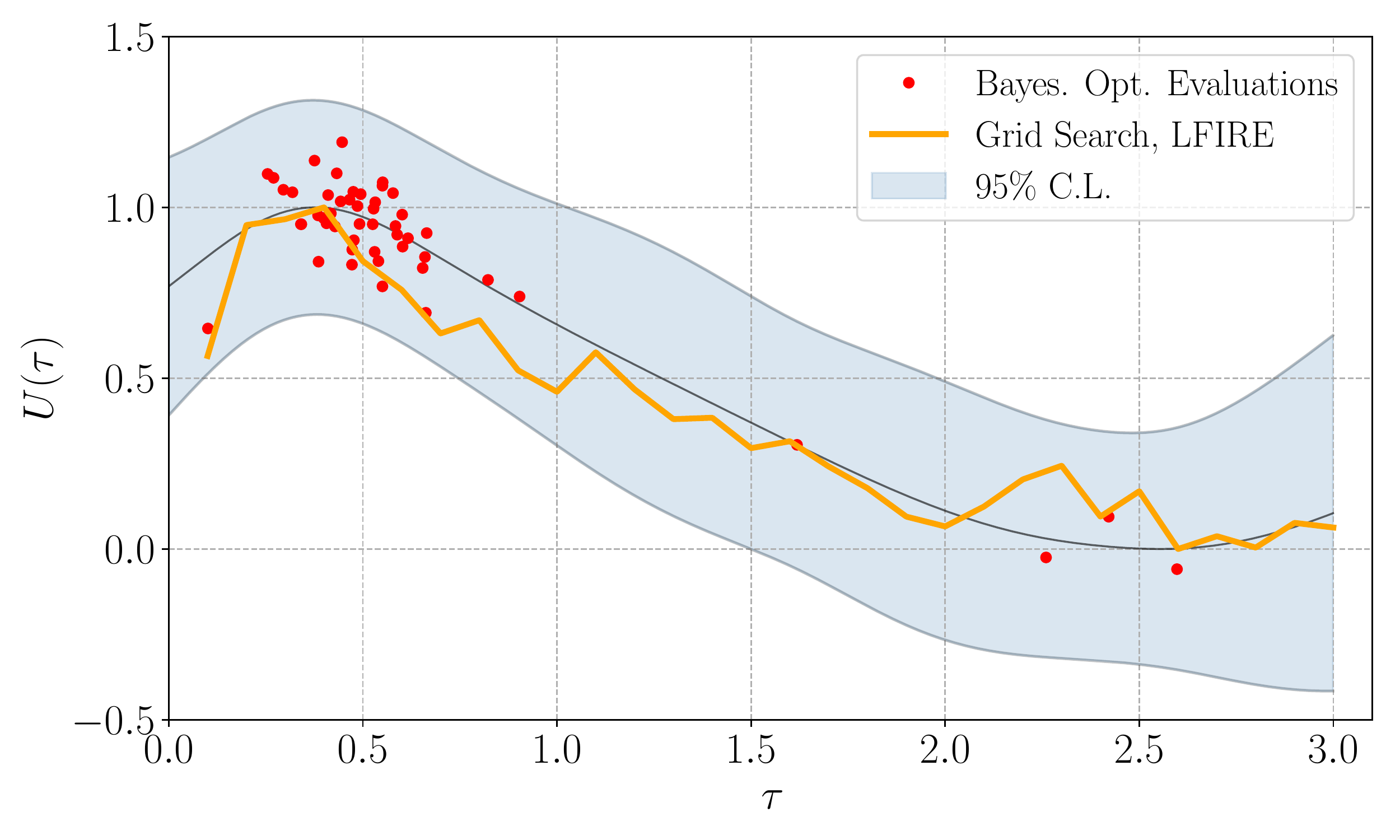}}
\caption[Filler]{Expected utilities $U(\tau)$ of the SIR Model for the grid search and BO method, using the design framework from Section~\ref{sec:prop}.}
\label{fig:sir_comparison}
\end{center}
\vskip -5mm
\end{figure}

We optimise the expected utility function according to the design framework outlined in Section~\ref{sec:prop}, using both grid search and Bayesian optimisation (BO). The resulting expected utility functions are shown in Figure~\ref{fig:sir_comparison}. Note that, unlike for the Death Model, we do not have a tractable likelihood function for the SIR Model and therefore we cannot compute an analytic expected utility function as a comparison.

The SIR Model expected utilities show a similar uni-modal behaviour as those of the Death Model, with the only difference being that the SIR Model results in a $U(\tau)$ with a peak that is more shifted towards lower $\tau$. The grid search and BO methods yield expected utilities that are generally similar to each other, with optimal design times $\tau^\ast=0.40$ and $\tau^\ast=0.44$, respectively. The $U(\tau)$ computed via grid search, however, has less resolution around the peak and generally more fluctuations due to the Monte Carlo approximation.

\begin{figure}%
\centering
\subfigure[Grid Search]{%
\label{fig:sirpost_first}%
\includegraphics[height=1.45in]{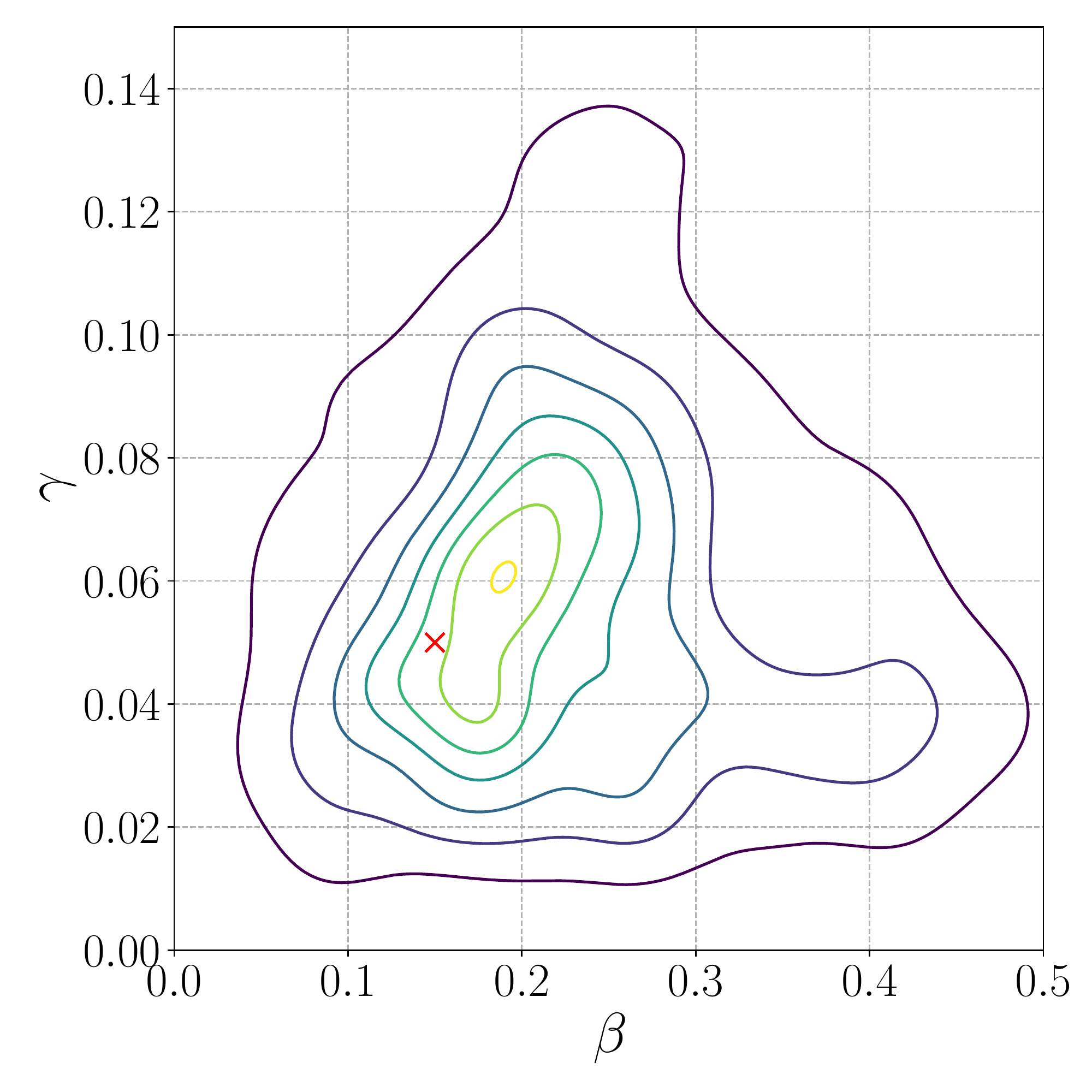}}%
\qquad
\subfigure[Bayes. Opt.]{%
\label{fig:sirpost_second}%
\includegraphics[height=1.45in]{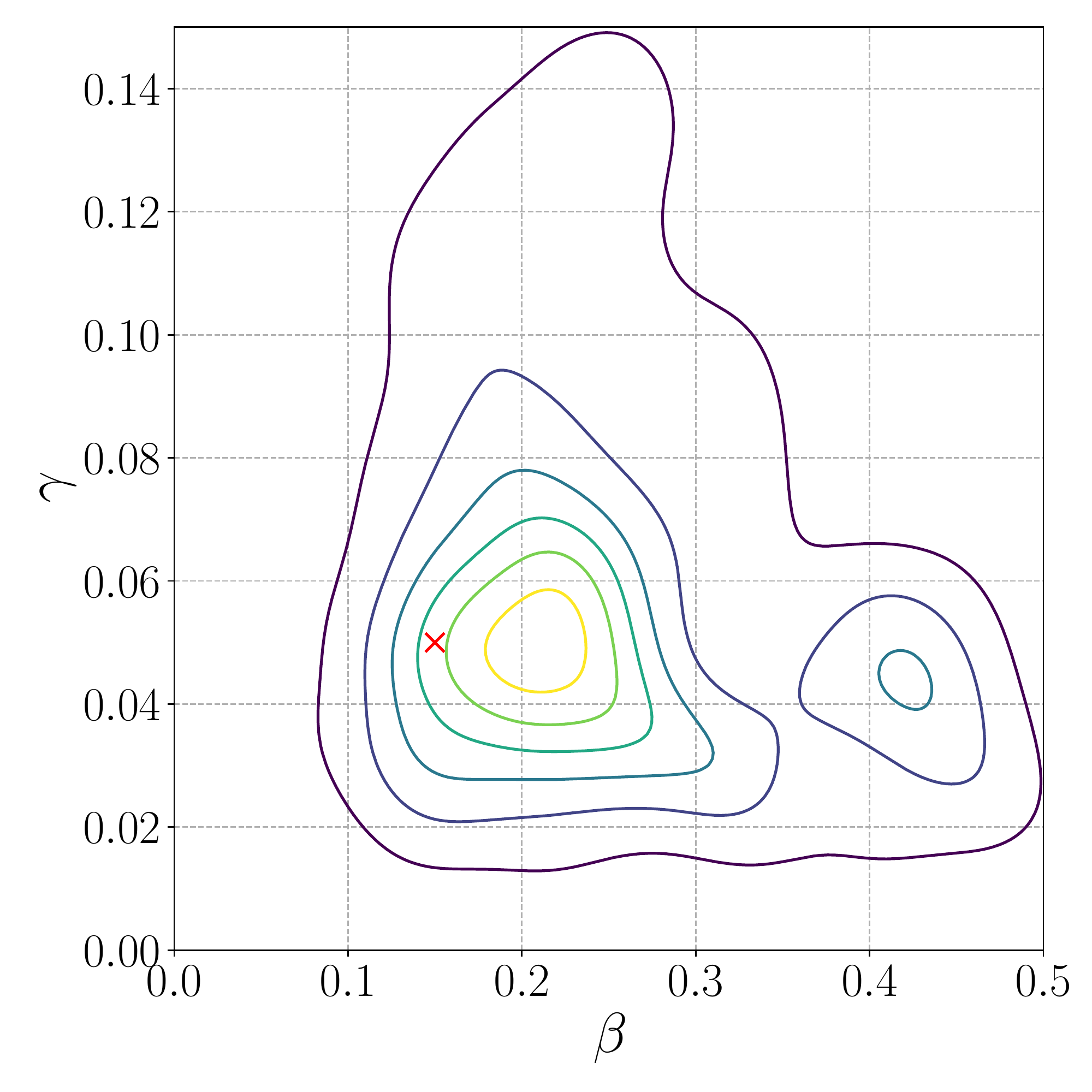}}%
\caption{SIR Model mean posterior densities for the grid search and BO method. The true parameter values are shown by a red cross.}
\end{figure}

After having found the optimal design points $\tau^\ast$ from the expected utilities in Figure~\ref{fig:sir_comparison}, we generate a 'real-world' observation $(S^\ast, I^\ast, R^\ast)$ by using $\beta_{\mathrm{true}} = 0.15$ and $\gamma_{\mathrm{true}} = 0.05$. Using this data we then obtain $10,000$ samples from the posterior distribution, by means of the procedure explained in Section~\ref{sec:prop}; we do this for both the grid search and BO method. We again apply Gaussian KDE to these samples to smooth out the posterior densities, and repeat this process for $50$ real-world observations. The mean posterior densities are shown in Figures~\ref{fig:sirpost_first} and~\ref{fig:sirpost_second}.

Both posterior distributions show a wide spread in $\beta$, a narrow spread in $\gamma$ and have uni-modal peaks that are in the same region.
The median parameter estimates when using the grid search method are $(\widehat{\beta}, \widehat{\gamma}) = (0.16, 0.05)$, and the 95\% credibility intervals are $(0.02, 0.23)$ for $\widehat{\beta}$ and $(0.01, 0.08)$ for $\widehat{\gamma}$. For the BO method we obtain $(\widehat{\beta}, \widehat{\gamma}) = (0.17, 0.04)$, and 95\% credibility intervals equal to $(0.03, 0.24)$ for $\widehat{\beta}$ and $(0.01, 0.07)$ for $\widehat{\gamma}$, both containing the true data generating parameters. Generally, the $\gamma$ parameter is well-estimated, whereas there is a marginal uncertainty in estimating the $\beta$ parameter.

\paragraph{High-Dimensional Designs}

As done for the Death Model, we shall now consider non-myopic design for the SIR Model, i.e.~situations where we know that we can take $n$ measurements. As before, the design variable becomes $n$-dimensional, $\mathbf{d} = [\tau_1, \tau_2, \dots, \tau_n]^\top$, with a similar constraint that the time must be ordered. The data vector is built from the observations $S(\tau_k)$, $I(\tau_k)$ and $R(\tau_k)$ at each time $\tau_k$, i.e.~$\mathbf{y} = [S(\tau_1), I(\tau_1), R(\tau_1), \dots, S(\tau_n), I(\tau_n), R(\tau_n)]^\top$. The computation of the LFIRE ratios is then done as previously, with the exception that the design variable is $n$-dimensional and the data vector is $3n$-dimensional. We again consider eight design dimensions; the expected utility for $n=8$ is evaluated as outlined in Algorithm~\ref{algo:mutual} and then optimised via Bayesian optimisation (BO), instead of grid search which is infeasible for $n=8$.

\begin{figure}%
\centering
\subfigure[Convergence]{%
\label{fig:multisir_first}%
\includegraphics[height=1.4in]{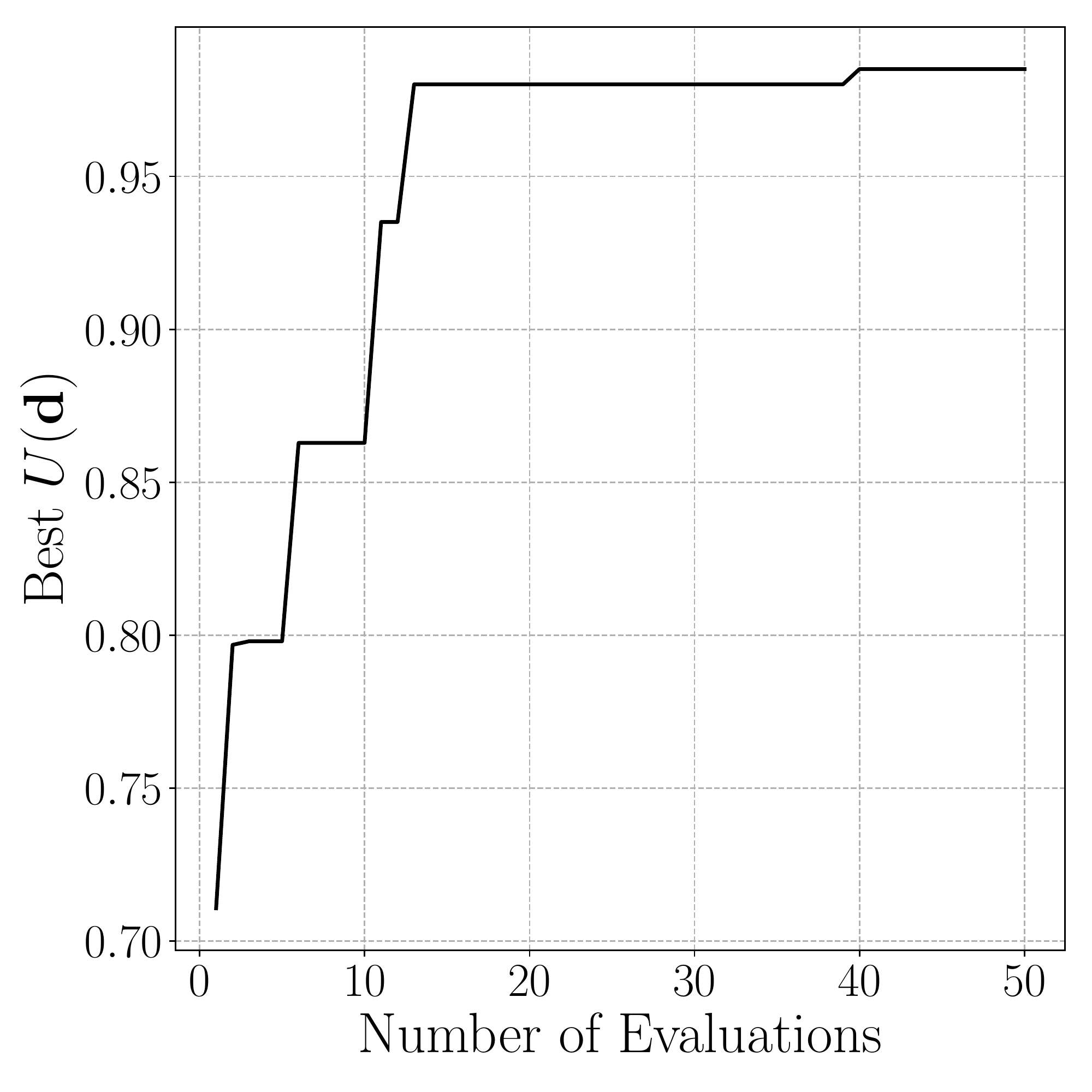}}%
\qquad
\subfigure[Posterior]{%
\label{fig:multisir_second}%
\includegraphics[height=1.4in]{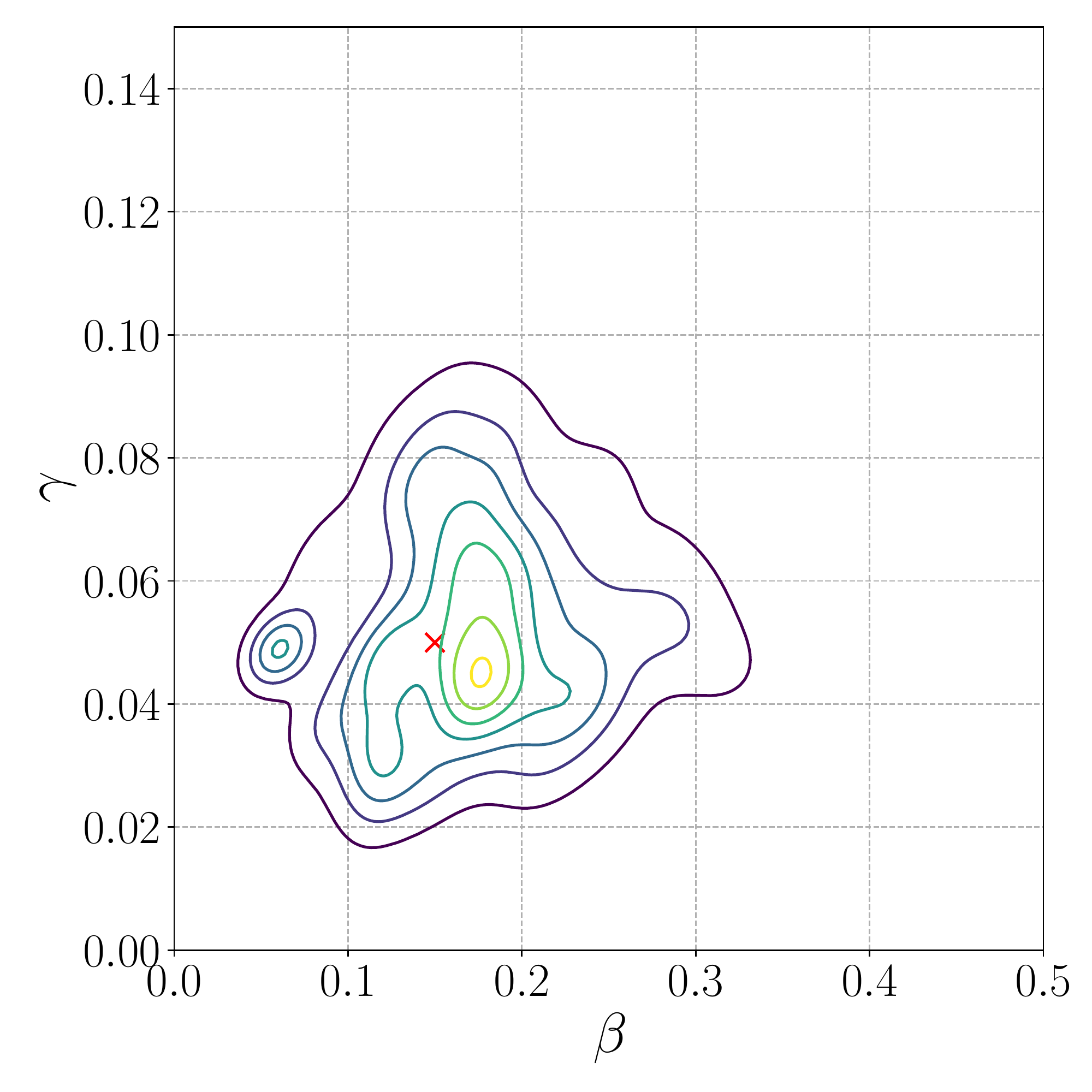}}%
\caption{SIR Model with $n=8$ design dimensions. (a) Maximal utility $U(\mathbf{d})$ identified as a function of number of evaluations. (b) The posterior distribution obtained with the optimal design $\mathbf{d}^\ast$, smoothed out by Gaussian kernel density estimation.}
\vskip -3mm
\end{figure}

Figure~\ref{fig:multisir_first} shows that, using BO, we can converge to the optimum in around $15$ expected utility evaluations. This is again a drastic difference to the number of evaluations we would have had to do with grid search.
We then generate $10{,}000$ LFIRE posterior samples at this optimal design, according to the procedure outlined in Section~\ref{sec:prop}, and smooth these out via Gaussian KDE. After repeating this process $50$ times, the resulting mean posterior density is shown in Figure~\ref{fig:multisir_second}, yielding parameter estimates (posterior means) $(\widehat{\beta}, \widehat{\gamma}) = (0.13, 0.05)$ and 95\% credibility intervals equal to $(0.02,0.18)$ for $\widehat{\beta}$ and $(0.02,0.07)$ for $\widehat{\gamma}$, both containing the true data generating parameters which were $\beta_{\mathrm{true}} = 0.15$ and $\gamma_{\mathrm{true}} = 0.05$.

We again compute the expected utility value at a design point $\mathbf{d}_{\mathrm{eq}}$ consisting of $8$ equidistant times, in order to compare it to the optimal design times. Using Algorithm~\ref{algo:mutual}, we obtain $U(\mathbf{d}_{\mathrm{eq}}) = 1.08$ for the equidistant design times and $U(\mathbf{d}^\ast) = 1.10$ for the optimal design times. While the optimal design has higher utility, the difference is small, which is again natural since the relative value of designing experiments generally diminishes as the amount of data that can be gathered increases.

\section{CONCLUSIONS} \label{sec:concl}

\vspace{-1ex}
In this paper, we have presented a Bayesian experimental design framework for implicit models, where the likelihood is intractable but sampling from the model is possible. We used the LFIRE approach to obtain density ratios of the posterior to prior distributions, which would otherwise not easily be possible with traditional likelihood-free inference methods such as approximate Bayesian computation. This allowed us to conveniently compute the mutual information between model parameters and simulated data, a notoriously difficult task for intractable models. We then used this mutual information as a utility function to decide where we should take data next. We optimised the expected utility by Bayesian Optimisation and found that this allowed us to find optimal designs in design dimensions impossible with grid search. An additional advantage of this approach is that it smoothes out the noise introduced by Monte-Carlo approximations.


There are a few limitations to our proposed design framework. First, high-dimensional Bayesian optimisation is still an active research area and its applicability in hundreds of dimensions remains to be investigated. Secondly, as with all likelihood-free inference methods, posterior estimations are approximate. We particularly noticed this when applying LFIRE to the Death Model. While the resulting utility functions were very similar and the optimal designs barely affected, characterising more generally how the approximation affects mutual information would be informative.

While we applied our framework to examples from epidemiology, the proposed methodology is general and thus applicable to a wider range of models. Other implicit models from neurobiology, cell biology or physics might be of particular interest, including both temporal and spatial models. It would then also be valuable to consider the cost or time required for doing an experiment and not only the information gain.

Finally, preliminary results suggest that the proposed framework extends to sequential designs where we update our belief about the model parameters based on the experimental outcome. This is a more realistic setting, but has barely been touched upon for models with intractable likelihoods.

\subsubsection*{Acknowledgements}
Steven Kleinegesse was supported in part by the EPSRC Centre for Doctoral Training in Data Science, funded by the UK Engineering and Physical Sciences Research Council (grant EP/L016427/1) and the University of Edinburgh.

\bibliography{references}


\end{document}